%% file: IEEEexample.tex
\documentclass[letterpaper, 10 pt, conference]{ieeeconf}  % Comment this line out if you need a4paper

\IEEEoverridecommandlockouts                              % This command is only needed if 
\usepackage{graphics} % for pdf, bitmapped graphics files
\usepackage{epsfig} % for postscript graphics files
\usepackage{times} % assumes new font selection scheme installed
\usepackage{amsmath} % assumes amsmath package installed
\usepackage{amssymb}  % assumes amsmath package installed
\usepackage{caption}
\usepackage{lipsum}
\usepackage{float}

\usepackage{algorithm}
\usepackage{algorithmic}
\usepackage{color}
\usepackage{booktabs}
\usepackage{subfigure}
\usepackage{setspace}
\usepackage{bm}
\usepackage{color}
\usepackage{multirow}
\usepackage{lipsum}
\usepackage{xcolor}
\usepackage{url}
\usepackage{hyperref}

\newcommand{\rebuttal}{\textcolor{black}}

\title{\LARGE \bf
DoReMi: Grounding Language Model by 
Detecting and Recovering from Plan-Execution Misalignment
}
\author{Yanjiang Guo$^{13*}$, Yen-Jen Wang$^{13*}$, Lihan Zha$^{2*}$, Jianyu Chen$^{13\dagger}$ \\ Project Page: \href{https://sites.google.com/view/doremi-paper}{https://sites.google.com/view/doremi-paper}
}

\begin{document}

\twocolumn[{%
\renewcommand\twocolumn[1][]{#1}%
\maketitle
\begin{center}
    \centering
    \vspace*{-5mm} 
    \includegraphics[width=1.00\linewidth]{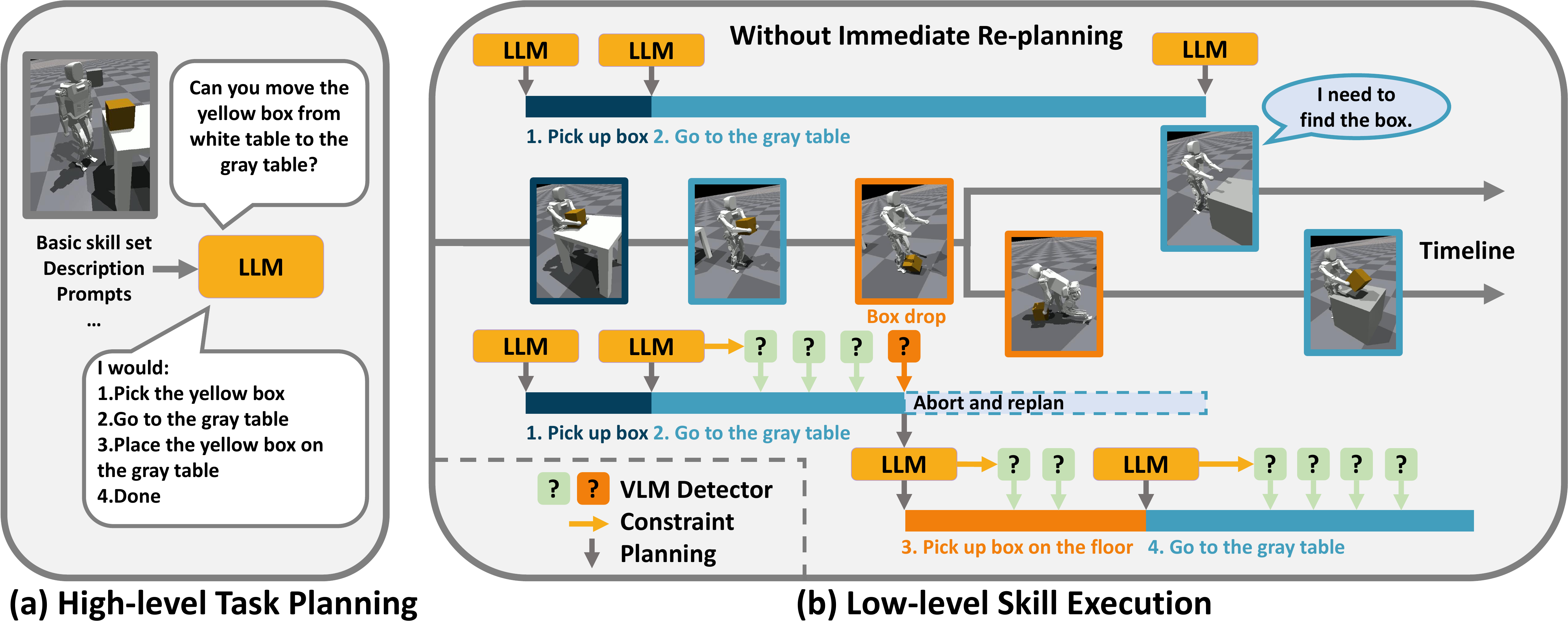}
    \vspace*{-3mm}
    % \captionof{figure}{Illustration of our motivation. Low-level execution may deviate from the high-level plan. DoReMi can immediately detect the misalignment between the plan and execution when the box drops accidentally and quickly recovers. Agents without immediate re-planning suffer from such misalignment.}
    \captionof{figure}{Illustration of our motivation. Previous works use LLM to generate only high-level textual plans. Therefore, Low-level execution may deviate from the high-level plan. We leverage LLM to generate both plans and constraints, which enables quick recovers when misalignments happen (e.g., box drop).}
    \label{motivation}
\end{center}%
}]

% \maketitle
\thispagestyle{empty}
\pagestyle{empty}

%%%%%%%%%%%%%%%%%%%%%%%%%%%%%%%%%%%%%%%%%%%%%%%%%%%%%%%%%%%%%%%%%%%%%%%%%%%%%%%%
\begin{abstract}

Large language models (LLMs) encode a vast amount of semantic knowledge and possess remarkable understanding and reasoning capabilities. Previous work has explored how to ground LLMs in robotic tasks to generate feasible and executable textual plans. However, low-level execution in the physical world may deviate from the high-level textual plan due to environmental perturbations or imperfect controller design. In this paper, we propose \textbf{DoReMi}, a novel language model grounding framework that enables immediate Detection and Recovery from Misalignments between plan and execution. Specifically, we leverage LLMs to play a dual role, aiding not only in high-level planning but also generating constraints that can indicate misalignment during execution. Then vision language models (VLMs) are utilized to detect constraint violations continuously. Our pipeline can monitor the low-level execution and enable timely recovery if certain plan-execution misalignment occurs. Experiments on various complex tasks including robot arms and humanoid robots demonstrate that our method can lead to higher task success rates and shorter task completion times. 

\end{abstract}

{
\renewcommand{\thefootnote}
{\fnsymbol{footnote}}
\footnotetext{$^*$Equal contribution, listed alphabetically.}

\renewcommand{\thefootnote}
{\fnsymbol{footnote}}
\footnotetext{$^\dagger$Corresponding author.{\tt\footnotesize jianyuchen@tsinghua.edu.cn}}

\renewcommand{\thefootnote}1
{\fnsymbol{footnote}}
\footnotetext[1]{Institute for Interdisciplinary Information Sciences, Tsinghua University, Beijing, China. {\tt\footnotesize guoyj22@mails.tsinghua.edu.cn}}

\renewcommand{\thefootnote}2
\footnotetext[2]{Weiyang College, Tsinghua University, Beijing, China.}

\renewcommand{\thefootnote}3
\footnotetext[2]{Shanghai Qi Zhi Institute, Shanghai, China.}
}
%%%%%%%%%%%%%%%%%%%%%%%%%%%%%%%%%%%%%%%%%%%%%%%%%%%%%%%%%%%%%%%%%%%%%%%%%%%%%%%%
\input{1intro_rela}

\input{2method}

\input{3exp}

\addtolength{\textheight}{-0cm}   % This command serves to balance the column lengths
                                  on the last page of the document manually. It shortens
                                  the textheight of the last page by a suitable amount.
                                  This command does not take effect until the next page
                                  so it should come on the page before the last. Make
                                  sure that you do not shorten the textheight too much.

%%%%%%%%%%%%%%%%%%%%%%%%%%%%%%%%%%%%%%%%%%%%%%%%%%%%%%%%%%%%%%%%%%%%%%%%%%%%%%%%

%%%%%%%%%%%%%%%%%%%%%%%%%%%%%%%%%%%%%%%%%%%%%%%%%%%%%%%%%%%%%%%%%%%%%%%%%%%%%%%%

%%%%%%%%%%%%%%%%%%%%%%%%%%%%%%%%%%%%%%%%%%%%%%%%%%%%%%%%%%%%%%%%%%%%%%%%%%%%%%%%
\section*{APPENDIX}

\subsection{Effectiveness of VLM finetuning}
After fine-tuning, the accuracy of VLM in the prepare-food task has significantly increased, as shown in Table \ref{accuray_hum}. We use the LoRA (Low-Rank adaptation)\cite{hu2021lora} method to finetune the BLIP-2 Flan-T5-xl model, the whole training process is finished on a single Nvidia A100 card.

\begin{table}[ht]
\small
\centering
\begin{tabular}{c|cccc|cccc}
\toprule[1.0pt]
 & \multicolumn{4}{c|}{Before finetune} & \multicolumn{4}{c}{After finetune} \\
        & TP & \rebuttal{FN} & FP & \rebuttal{TN} & TP & \rebuttal{FN} & FP & \rebuttal{TN} \\ \midrule[0.5pt]
Obstacle & 120 & 5 & 0 & 14 & 121 & 4 & 0 & 14 \\ 
Move box & 140 & 0 & 6 & 22 & 140 & 0 & 2 & 26 \\ 
Prepare food & 78 & 27 & 8 & 25 & 99 & 6 & 1 & 32 \\ \bottomrule[1.0pt]
% Obstacle & 120 & 5 & 0 & 14 & 121 & 4 & 0 & 14 \\ 
% Move box & 140 & 0 & 6 & 22 & 140 & 0 & 2 & 26 \\ 
% Prepare food & 78 & 27 & 8 & 25 & 99 & 6 & 1 & 32 \\ \bottomrule[1.0pt]
\end{tabular}
\caption{Accuracy analysis of VLM on humanoid tasks.}\label{accuray_hum}
\end{table}

\subsection{VLM ablation study}
We also conduct an ablation study on different types of VLM with max disturbances in our tasks, as shown in Table \ref{ablation_vlm}. The BLIP-2 model performs similarly to the Instruct-BLIP model. However, all zero-shot transferred models can not perform well in complicated prepare-food tasks.
\begin{table}[ht]
\centering
\begin{tabular}{cccc}
\toprule[1.0pt]
\textbf{Success rate\%} & \textbf{BLIP-1} & \textbf{BLIP-2} & \textbf{Instruct-BLIP} \\ \midrule[0.5pt]
Obstacle-avoidance & 88 & 90 & 92 \\
Move-box & 64 & 94 & 92 \\
Prepare-food & 16 & 37 & 40\\
\bottomrule[1.0pt]
\end{tabular}
\caption{Ablation study on zero-shot transferred VLM}\label{ablation_vlm}
\end{table}

\subsection{Detalied pipeline}
\textbf{Example for stack-in-order task:}

\textbf{LLM input:} The robot performs manipulation tasks. At the same time, the robot needs to satisfy some constraints to ensure the successful execution of each task. Just fill in the blank and directly output the next step. Task: Stack blocks in the order of brown, red, and green.

\textbf{LLM output step by step:} (1) Pick the red block, [Constraint: The robot is holding red block], 

(2) Place the red block on the brown block, [Constraint: The red block is on the brown block], 

(3) Pick the green block, [Constraint: The robot is holding green block, the red block is on the brown block], 

(4) Place the green block on the red block, [Constraint: The red block is on the brown block, the green block is on the red block], \textbf{[Detector feedback: the red block is not on the brown block]}, 

(5) Pick the green block, [Constraint: The robot is holding green block, the red block is on the brown block], 

(6) Place the red block on the brown block, [Constraint: The red block is on the brown block], 

(7) Pick the green block, [Constraint: The robot is holding green block, The red block is on the brown block], 

(8) Place the green block on the red block, [Constraint: The red block is on the brown block, The green block is on the red block]

% \section*{ACKNOWLEDGMENT}

% The preferred spelling of the word ÒacknowledgmentÓ in America is without an ÒeÓ after the ÒgÓ. Avoid the stilted expression, ÒOne of us (R. B. G.) thanks . . .Ó  Instead, try ÒR. B. G. thanksÓ. Put sponsor acknowledgments in the unnumbered footnote on the first page.

%%%%%%%%%%%%%%%%%%%%%%%%%%%%%%%%%%%%%%%%%%%%%%%%%%%%%%%%%%%%%%%%%%%%%%%%%%%%%%%%

% References are important to the reader; therefore, each citation must be complete and correct. If at all possible, references should be commonly available publications.

\bibliographystyle{IEEEtran}
\bibliography{IEEEexample}

\end{document}

%% file: 1intro_rela.tex
\section{Introduction}

% \wyj{第一段：凸显LLM和robotics的结合，不要直接从decision-making出发}
% \wyj{第二段：除了low-level skill外，还要要包括language-action，language到control有两种方法，要解释为什么用low-level skill, such as dynamics 很复杂的 1. 计算频率太低(robo transformer 3hz?) 2. 本质上是behavior cloning。我们关注的是grounding}
% \wyj{第三段： grounding的问题- 要能执行好，上下层feedback的同步。先强调我们做的点：上下层对齐或什么，previous works虽然也有靠虑feedback，但他只有衔接时候才有。先强调我们做的点：上下层对齐或什么，previous works虽然也有靠虑feedback，但他只有衔接时候才有。举几个例子，说之前的不行。}
 
Large language models (LLMs) pre-trained on web-scale data emerge with common-sense reasoning ability and understanding of the physical world. Previous works have incorporated language models into robotic tasks to help embodied agents better understand and interact with the world to complete challenging long-horizon tasks that require complex planning and reasoning \cite{ahn2022can,huang2022language,liang2022code}. 

To make the generated plan executable by embodied agents, we need to ground the language. 
One line of the works leverages pre-trained language models in an end-to-end manner that directly maps language and image inputs to the robot's low-level action space \cite{brohan2022rt,brohan2023rt, jang2022bc,shridhar2023perceiver,nair2022learning}.
These approaches often require large amounts of robot action data for successful end-to-end training, which is expensive to acquire \cite{brohan2022rt}. Moreover, these action-output models often contain large transformer-based architectures and cannot run at high frequencies. Therefore, they may not be suitable for tasks with complex dynamics (e.g., legged robots) that require high-frequency rapid response. Recently, many works have adopted a hierarchical approach where language models perform high-level task planning, and then some low-level controllers are adopted to generate the complex robot control commands \cite{ahn2022can,huang2022language,liang2022code,huang2022inner}. Under this hierarchical framework, we can leverage powerful robot control methods, such as reinforcement learning, to handle complex robot dynamic control problems with high frequency.

However, these grounding methods often assume that every low-level skill can perfectly execute the high-level plan generated by the language model. 
In practice, low-level execution may deviate from the high-level plan due to environmental perturbations or imperfect controller design. These misalignments between plan and execution may occur at any time during the task procedure. Previous works consider incorporating execution feedback into language prompts once the previous plan step is finished. If the step is unsuccessful, the process is repeated \cite{huang2022inner}.
However, this delayed feedback can be inefficient.
For instance, as illustrated in Figure \ref{motivation}(b), when a human is carrying a box and performing the low-level skill ``Go to the gray table", if the box is accidentally dropped, it becomes futile to continue with the current skill. The human will immediately abort the current skill and call for the skill ``Pick up the box". However, agents without immediate re-planning will continue going forward and will take more time to pick up the box dropped halfway after reaching the destination.

% \begin{figure}[ht]
%     \centering
%     \includegraphics[width=0.95\textwidth]{figures/pic1_clip.pdf}
%     % \vspace{0.2mm}
%     \caption{Illustration of our motivation. Low-level execution may deviate from the high-level plan. DoReMi can immediately detect the misalignment between the plan and execution when the box drops accidentally and quickly recovers. Agents without immediate re-planning suffer from such misalignment.}
%     \label{motivation}
%     \vspace{-1mm}
% \end{figure}
In this paper, we propose a novel framework \textbf{DoReMi} which enables immediate \textbf{D}etecti\textbf{o}n and \textbf{Re}covery from plan-execution \textbf{Mi}salignments. 
Specifically, in addition to employing LLMs for high-level planning \cite{ahn2022can}, we further leverage LLMs to generate constraints for low-level execution based on their understanding of physical worlds. 
During the execution of low-level skills, a vision language model (VLM) \cite{li2023blip} is employed as a general "constraint detector" to monitor whether the agent violates any constraints continuously. If some constraints are violated, indicating that the plan and execution may be misaligned, the language model is immediately called to re-plan for timely recovery. 
% We summarize several advantages of our pipeline: (1) LLM plays a dual role, aiding not only in high-level planning but also in supervising low-level execution, enabling rapid detection and recovery; (2) The VLM can focus on the specific constraints suggested by the LLM and only need to pick binary answers, providing more precise feedback. This collaborative approach between the LLM and the VLM can help align the plan and execution during the whole task period. 
% Experiments in physical simulations, including robot arm manipulation tasks and humanoid robot tasks, demonstrate that DoReMi leads to a higher task success rate and shorter task execution time.
Our contributions can be summarized as follows:
\begin{itemize}
    \item Different from previous works that use LLM only to plan, we leverage LLM to play a dual role, aiding not only in high-level planning but also generating constraints to supervise low-level execution.
    \item We propose DoReMi, an integrated framework between LLMs and VLMs to enable more precise and frequent feedback automatically.
    \item Experiments on robot arm manipulation tasks and humanoid robot tasks demonstrate that DoReMi leads to a higher task success rate and shorter task execution time.
\end{itemize}

% only need to pick a binary answer
%第三点不用
% Our contributions can be summarized as follows:
% \begin{itemize}
%     \item We show the importance of aligning high-level plan and low-level execution and propose the DoReMi framework, which enables immediate detection and recovery from plan-execution misalignments for complex long-horizon robotic problems.
%     \item We leverage LLMs for both planning and constraint generation. During low-level skill execution, we use VLMs as a general constraint detector to timely detect plan-execution misalignment and re-plan.
%     \item Theoretical analyses and experiments on various complex robotic tasks verify the effectiveness of DoReMi, which results in less task execution time and higher success rates.
% \end{itemize}

\section{Related Works}

\textbf{Language Grounding}
Prior research has attempted to employ language as task abstractions and acquired control policies that are conditioned on language \cite{macmahon2006walk,chaplot2018gated,jiang2019language,misra2017mapping,mei2016listen}. 
Furthermore, some studies have investigated the integration of language and vision inputs within embodied tasks to directly predict the control commands \cite{silva2021lancon,guhur2023instruction,goyal2021pixl2r}. 
Recent works, including \cite{brohan2022rt,brohan2023rt,shridhar2023perceiver,zhang2021hierarchical,lynch2022interactive}, have demonstrated significant progress in utilizing transformer-based policies to predict actions. However, these end-to-end approaches heavily depend on the scale of expert demonstrations for model training.

\textbf{Task Planning with Language Model}
Traditionally, task planning was solved through symbolic reasoning \cite{nau1999shop,fikes1971strips} or rule-based planners \cite{fox2003pddl2, jiang2019task}. Recently, many works demonstrated that large language models (LLMs) can generate executable plans in a zero/few-shot manner with appropriate grounding \cite{huang2022language,ahn2022can,zeng2022socratic,ren2023robots}. Some pre-trained low-level skills (primitives) are then adopted to execute steps in order. These LLM planners typically assume the successful execution of each skill, resulting in an open-loop system in physical worlds. Works in the instruction-following benchmark \cite{shridhar2020alfred,puig2018virtualhome} like ReAct \cite{yao2022react}, and Reflexion \cite{shinn2023reflexion}, incorporate feedback into LLM prompts to help planning after each step of the plan is finished. However, these benchmarks operate in discrete scenes and pay less attention to the skill execution period. The closest work to ours is Inner Monologue \cite{huang2022inner}, which also considers continuous physical worlds, and takes into account 3 types of feedback (e.g. success detectors, scene descriptions, and human feedback) upon the completion of each step. However, Inner Monologue needs manually designed queries to get information from environments, which is impractical and hard to obtain at high frequency. In contrast to this, \textit{our framework enables precise and high-frequency feedback with practical detectors automatically}.

\textbf{Vision Language Model for Embodied Control.}
The vision language model (VLM) is trained on image-text pairs, enabling it to simultaneously understand visual and textual inputs and address a variety of downstream tasks, such as visual question answering (VQA)\cite{li2023blip,antol2015vqa}, image captioning \cite{zhou2020unified}, and object detection \cite{gu2021open}. VLMs align semantic information between vision and natural language, thereby aiding in grounding language models and facilitating embodied control. 
Pre-trained visual encoders or instruction encoders \cite{radford2021learning} can be connected with some action head to help train end-to-end policies \cite{shridhar2022cliport} or generate textual plans \cite{driess2023palm}. RT-2 \cite{brohan2023rt} directly fine-tuned on a VLM can generate texts and robot control actions simultaneously. VLMs can also act as scene descriptors\cite{huang2022inner}, success detectors \cite{du2023vision,zhang2023grounding}, or object detectors\cite{stone2023open} to facilitate the task execution.
% Pretrained models such as CLIP \cite{radford2021learning} have been integrated into diverse embodied tasks \cite{khandelwal2022simple}. CLIPort \cite{shridhar2022cliport}, which incorporates Transporter \cite{zeng2021transporter} with CLIP, effectively combines spatial precision and semantic understanding. The PaLM-E model \cite{driess2023palm}, equipped with pretrained vision transformer \cite{dehghani2023scaling} and PaLM model \cite{chowdhery2022palm}, can leverage multimodal inputs and generate textual plans directly. Inner-Monologue \cite{huang2022inner} assumes perfect VLMs as success detectors and scene descriptors to obtain task feedback. 
To ensure adherence to crucial constraints, we employ the VLM \cite{li2022blip} as a "constraint detector", periodically verifying whether the agent satisfies specific constraints.

\begin{figure*}[ht]
% \vspace{-2mm}
\centering
\includegraphics[width=0.8\textwidth]{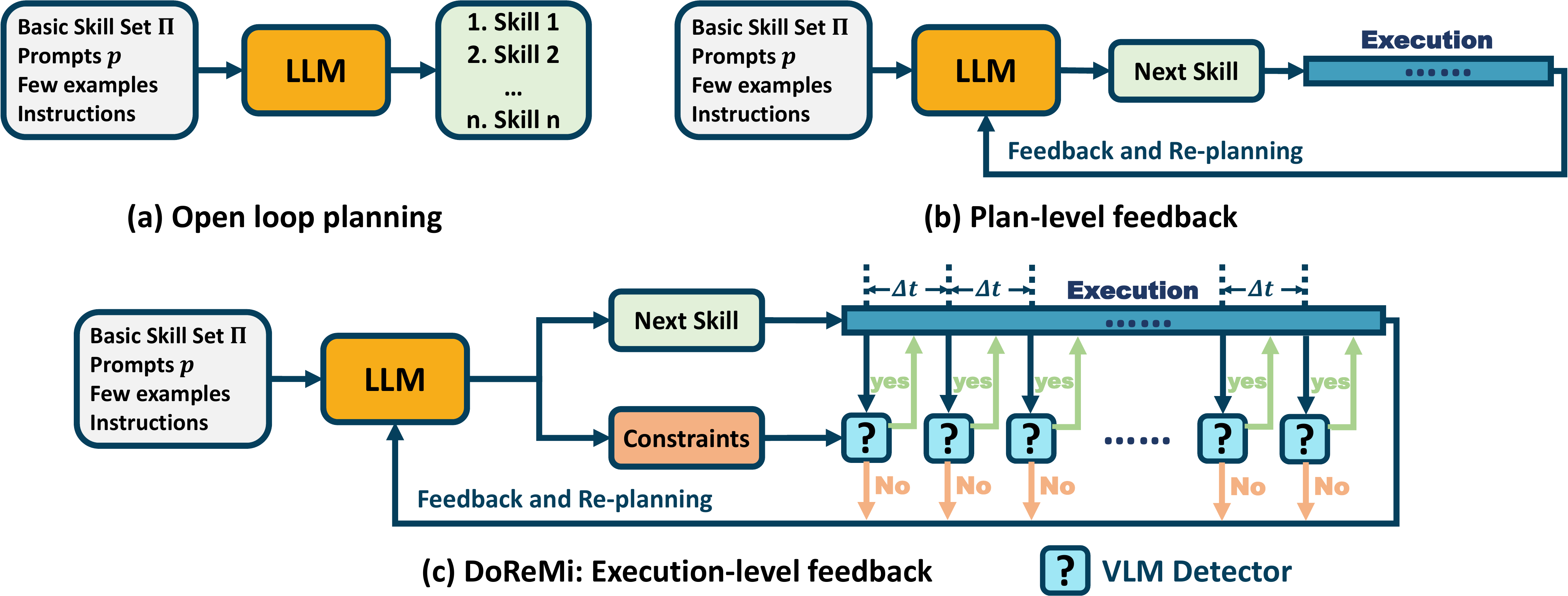}
\caption{Previous methods perform open-loop planning or only re-plan when the previous skill is finished. Our DoReMi framework leverages LLM to generate both the plan and corresponding constraints. Then a VLM is employed to supervise the low-level execution period, which enables immediate recovery from plan-execution misalignment.}
\label{method_new}
\vspace{-4mm}
\end{figure*}

\begin{figure*}[t]
\begin{center}
\subfigure[Open-ended scene description]{
\includegraphics[width=0.38\linewidth]{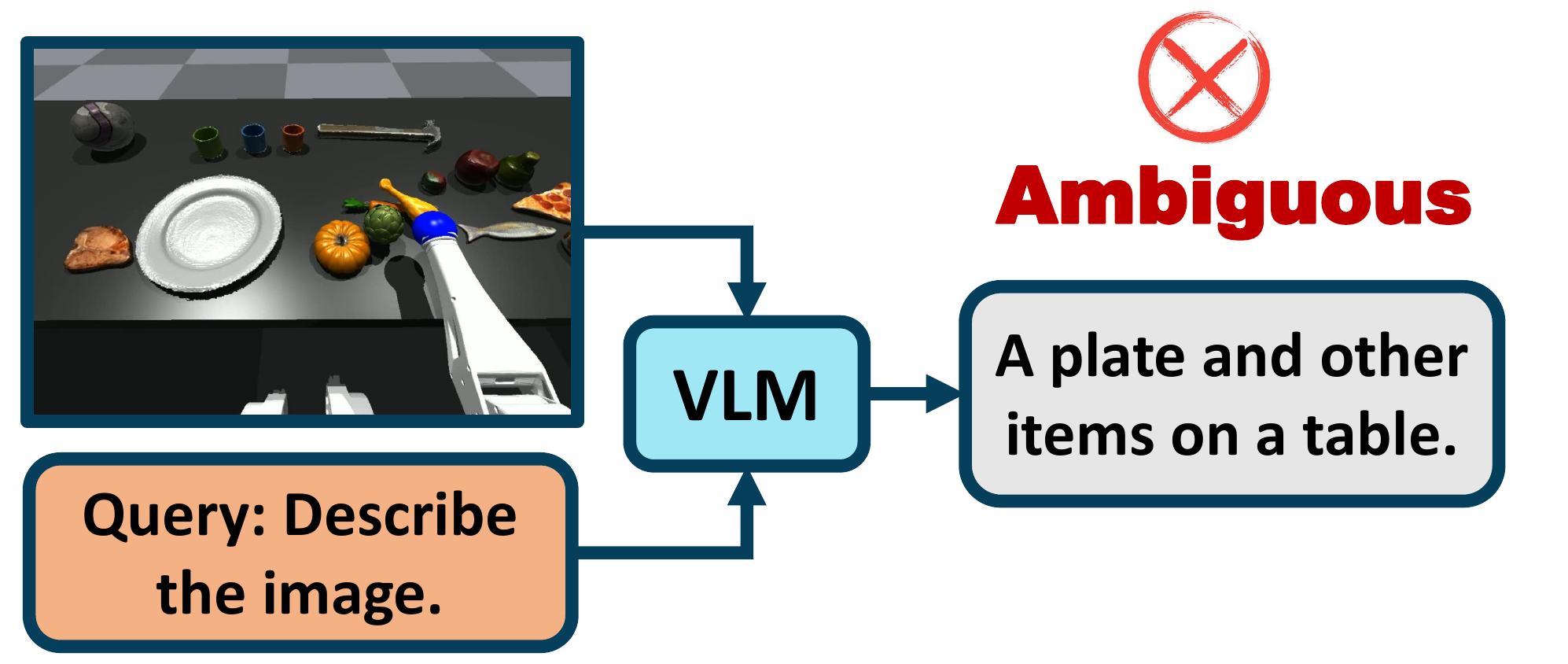}}
\subfigure[VLMs focus on specific constraint generated by LLMs]{
\includegraphics[width=0.58\linewidth]{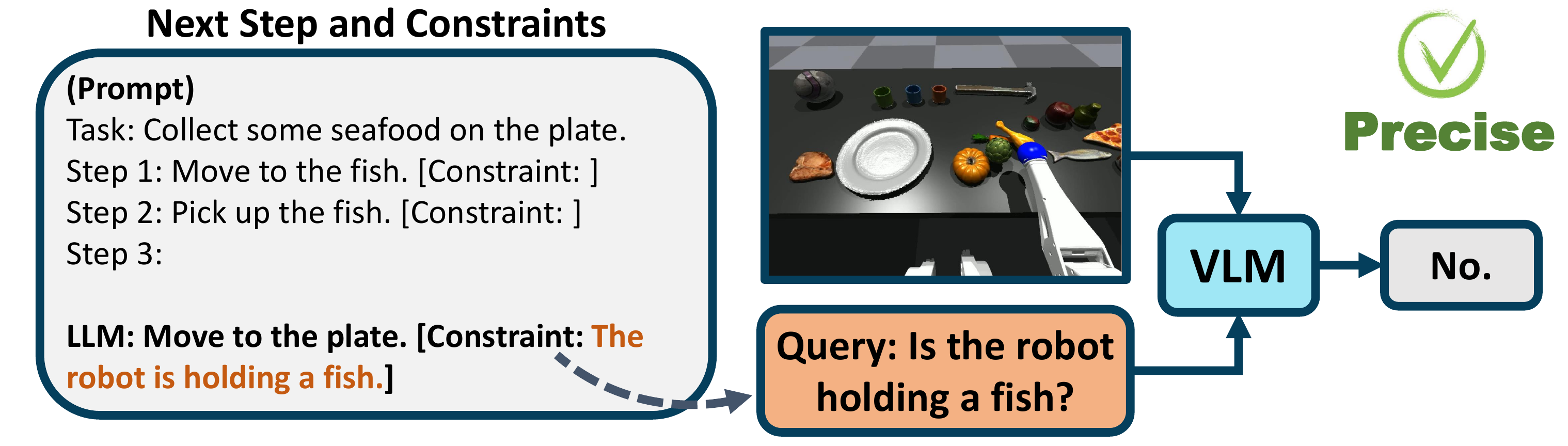}}
\vspace{-2mm}
\caption{Open-ended scene descriptions of VLMs are ambiguous. DoReMi leverages the LLM to reason specific constraints and actively queries the VLM for key information, resulting in much more precise feedback.}
\label{vlm_compare}
\end{center}
\vspace{-5mm}
\end{figure*}

%% file: 2method.tex
\section{Method}
In this section, we introduce our \textbf{DoReMi} framework which enables immediate \textbf{D}etecti\textbf{o}n and \textbf{Re}covery from Plan-Execution \textbf{Mi}salignment. Our algorithm can be succinctly described in two stages depicted in Figure \ref{method_new}(c):

\begin{enumerate}
    \item At the high-level planning stage, given a set of low-level skills, prompts, and high-level task instruction, language models are leveraged to play a dual role, aiding not only in planning the next skill but also generating constraints for the next skill based on historical information.
    \item During the low-level skill execution stage, we employ a vision-language model (VLM) \cite{li2023blip} as a general "constraint detector" that periodically verifies the satisfaction of all constraints. If any constraint is violated, the language model is invoked for immediate re-planning to facilitate recovery.
\end{enumerate}
\vspace{-2mm}
% Visualizations of the pipeline can be found in Figure \ref{method}. Next, we introduce these 3 components specifically.

\subsection{Language Model for Planning}
\rebuttal{Following previous works that leverage LLM to generate feasible textual plans\cite{ahn2022can}}, we utilize LLMs to plan the next steps through few-shot in-context learning. Furthermore, we employ language models for re-planning when our constraint detector identifies a plan-execution misalignment. In such scenarios, we additionally include the misalignment information in prompts and invoke the LLM for re-planning. 
% Detailed planning prompts can be found in Appendix \ref{prompt}. 
Practically, we deploy the Vicuna-13B model \cite{vicuna2023} locally and pick the next skill with max output probability. We also try GPT4 \cite{openai2023gpt} through OpenAI API to directly output the next step with zero temperature. Both LLMs exhibit effective planning capabilities in our tasks.

\subsection{Language Model for Constraint generation}\label{constraint_set}
LLM planner helps agents decompose long-horizon tasks into skill sequences. However, LLMs are not inherently integrated into the execution of low-level skills, which potentially leads to misalignment between plan and execution. To further explore the ability of LLMs in embodied tasks, we utilize LLMs not only for next-step planning but also for constraint generation based on historical information.  
For instance, consider the execution period of the \textit{``go to"} skill after the \textit{``pick up box"} skill. In such cases, the constraint \textit{``robot holds box"} must be satisfied and violation of this constraint could indicate a failure in the picking or possible dropping of the box. Similarly, after the skill \textit{``place red block on green block"}, the constraint \textit{``red block on green block"} should always be met.
LLMs possess the capability to automatically generate these constraints for planned steps, drawing upon their encoded understanding of the physical world. Moreover, the VLM detector can focus on these specific constraints and only needs to pick binary answers from ``Yes" or ``No", resulting in much more precise feedback. In contrast, open-ended scene descriptions of VLMs may result in large ambiguity and miss essential information, as shown in Figure \ref{vlm_compare}.

In practice, after the LLM selects the next step with the highest output probability, we continue the generation starting with ``Constraint:" to derive specific constraints. 

\begin{figure*}[t]
    \centering
    \includegraphics[width=1.0\linewidth]{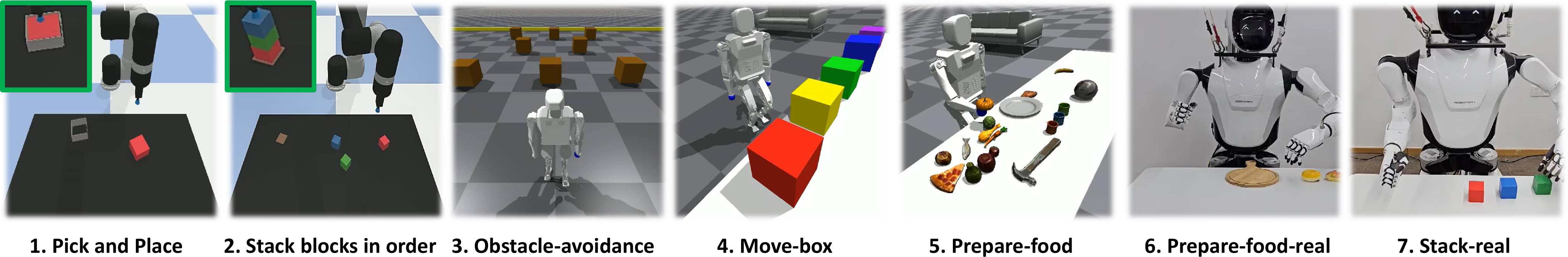}
    \caption{Robot manipulation and humanoid robot tasks in our experiments. We consider various types of environmental disturbance and imperfect controllers in both simulation and the real world.}
    \label{exp}
    \vspace{-4mm}
\end{figure*}
\begin{figure}[t]
\vspace{-2mm}
\begin{algorithm}[H]
   \caption{\textbf{DoReMi} (Immediate \textbf{D}etecti\textbf{o}n and \textbf{Re}covery from \textbf{Mi}salignment)}
   \label{algo1}
   \textbf{Given:} A high level instruction $i$, a skill set $\Pi$, language description $l_{\Pi}$ for $\Pi$, language model $L$, prompt $p_0$, and VLM constraint detector $D$.
\begin{algorithmic}[1]
   % \STATE{\textcolor[RGB]{100,100,220}{\# Initialize}}
   \STATE{Initialize the skill sequence $\pi \leftarrow \emptyset$, the number of steps $n \leftarrow 1$.}
   \WHILE{$l_{\pi_{n-1}}\neq done$}
   \STATE{$\pi_n \leftarrow \arg\max_{\pi\in\Pi}L(l_{\pi}|i,p_{n - 1},l_{\pi_{n-1}},..l_{\pi_{0}})$, $c_n \leftarrow L(i,p_{n-1}, l_{\pi_{n}},..l_{\pi_{0}})$}
   \STATE{Update prompt $p_n$.}
   \WHILE{$\pi_n$ is not finished}
   \STATE{Every $\Delta t$ second, query agent all the constraints $c_n$ using the constraint detector $D$.}
   \IF{$\exists D(c_n)= \FALSE$}
   \STATE{Add constraint violate information into prompt $p_n$ and \textbf{break}.}
   \ENDIF
   \ENDWHILE
   \STATE{$n \leftarrow n + 1$.}
   % \STATE{$n \leftarrow n + 1$}
   \ENDWHILE
\end{algorithmic}
\end{algorithm}
% \vspace*{-1.4cm}
\vspace{-9mm}
\end{figure}

\subsection{VLM as Constraint Detector}
% \vspace{-2mm}
% To check the constraints in a general manner without the need for specific designs or hard coding\gyj{ambiguous?}, we employ a vision language model(VLM) \cite{li2023blip} as a general "constraint detector", denoted as $D$. 
Subsequent to the constraint generation stage, the agent proceeds to execute the planned step while adhering to constraints suggested by the LLM. 
The LLM-generated constraints may include various types, such as "red block is on blue block," "no obstacles in front of the robot," "robot is holding an apple," and more.
In this work, we adopt a vision language model(VLM) \cite{li2023blip} as a general "constraint detector" to check all constraints through visual information. 
The visual input of the VLM is captured from either a first-person or third-person perspective camera, and the text input is automatically adapted from the LLM proposed constraints in the form "Question: Is the \textit{constraint $c_j$ satisfied}? Answer:". 
For each query, the VLM only needs to select an answer from \{``Yes'', ``No''\}, which consists of very short token lengths and costs less than 0.1 second. We use $D(c_j)$ to denote the answer of the VLM $D$ when checking constraint $c_j$. If $c_j$ is satisfied, $D(c_j) = True$; otherwise, $D(c_j) = False$. The pseudo-code of the pipeline is provided in Algorithm \ref{algo1}. It's also worth mentioning that detectors in other modalities are also compatible with our framework and constraint detectors can run parallel to low-level controllers with different frequencies.  
% \gyj{I have changed this part. Is it ok?}
% \The VLM detector can run at 5-10 Hz and the low-level controller can run at more than 50 Hz. \gyj{Other types of detectors?}

In practice, we use the pre-trained BLIP-2 model \cite{li2023blip} as a general "constraint detector" to periodically check whether the agent satisfies all constraints every $\Delta t=0.2$ second. If so, the robot continues executing the current low-level skill; otherwise, the robot aborts the current skill, and the re-planning process is triggered. We observe that pre-trained zero-shot VLM can perform well in most tasks, except those with extremely complex scenes. To enhance the performance in such complex tasks, we collect a small dataset and fine-tune the VLM using the parameter-efficient LoRA method \cite{hu2021lora}. We also verify that the fine-tuned VLM detector can generalize to unseen objects, unseen backgrounds, and even unseen tasks.

% For relatively simple tasks, we just use the zero-shot transferred BLIP-2 model. However, as tasks become complicated, the performance of the zero-shot VLM drops. So we collect several demonstrations and fine-tune the BLIP-2 model in our domain with parameter efficient LoRA method \cite{hu2021lora}. We also verify that fine-tuned VLM can generalize to unseen objects, unseen backgrounds, and even unseen tasks. 

% To summarize, our method has three main components: (1) We use language models for planning and constraint set update. (2) We adopt the VQA model to check constraint satisfaction during the whole execution period. (3) If any constraint is violated, we trigger timely re-planning with the language model. The pseudo-code is provided in Algorithm \ref{algo1}.

%% file: 3exp.tex
\begin{table*}[t]\centering
\small
% \tiny
\begin{tabular}{cl|ccccc|ccc}\toprule[1.0pt]
\multicolumn{2}{c|}{\multirow{2}{*}{\textbf{Tasks with disturbance}}} & \multicolumn{5}{c|}{\textbf{Success Rate(\%) \bm{$\uparrow$}}} & \multicolumn{3}{c}{\textbf{Execution Time(s) \bm{$\downarrow$}}} \\ 
\multicolumn{2}{c|}{} & \textbf{\begin{tabular}[c]{@{}c@{}}SayCan\end{tabular}} & \textbf{\begin{tabular}[c]{@{}c@{}}CLIPort $ $\end{tabular}} & \textbf{\begin{tabular}[c]{@{}c@{}}IM\end{tabular}} & \textbf{\begin{tabular}[c]{@{}c@{}}DoReMi\\ (ours)\end{tabular}} & \textcolor{gray}{\textbf{\begin{tabular}[c]{@{}c@{}}IM-\\ Oracle\end{tabular}}} & \textbf{\begin{tabular}[c]{@{}c@{}}IM\end{tabular}} & \textbf{\begin{tabular}[c]{@{}c@{}}DoReMi \\(ours)\end{tabular}}  & \textcolor{gray}{\textbf{\begin{tabular}[c]{@{}c@{}}IM-\\ Oracle\end{tabular}}} \\ \midrule[0.5pt]
\multirow{3}{*}{\textbf{\begin{tabular}[c]{@{}c@{}}Pick and place \\ with random drop $\bm{p}$\end{tabular}}} & \textbf{$\bm{p}$=0.0} &100\,($\pm$0)& 100\,($\pm$0)&100\,($\pm$0) & 100\,($\pm$0) & \textcolor{gray}{100\,($\pm$0)} & \textbf{2.7\,($\pm$0.0)} &\textbf{2.7\,($\pm$0.0)} &\textcolor{gray}{2.7\,($\pm$0.0)} \\
 & \textbf{$\bm{p}$=0.2} & 81\,($\pm$9)&100\,($\pm$0)& 100\,($\pm$0) & 100\,($\pm$0) & \textcolor{gray}{100\,($\pm$0)} & 3.4\,($\pm$0.2) & \textbf{3.0\,($\pm$0.2)} & \textcolor{gray}{3.4\,($\pm$0.2)} \\
 & \textbf{$\bm{p}$=0.3} &63\,($\pm$9)&100\,($\pm$0)& 100\,($\pm$0) & 100\,($\pm$0) & \textcolor{gray}{100\,($\pm$0)} & 4.0\,($\pm$0.2) &\textbf{3.3\,($\pm$0.2)} & \textcolor{gray}{4.0\,($\pm$0.2)} \\ \midrule
 
\multirow{4}{*}{\textbf{\begin{tabular}[c]{@{}c@{}}Stack in order\\ with noise $\bm{n}$\end{tabular}}} & \textbf{$\bm{n}$=0.0} &\textbf{100\,($\pm$0)}&\textbf{100\,($\pm$0)}& \textbf{100\,($\pm$0)} & \textbf{100\,($\pm$0)} & \textcolor{gray}{100\,($\pm$0)} & \textbf{7.2\,($\pm$0.0)} & \textbf{7.2\,($\pm$0.0)} & \textcolor{gray}{7.2\,($\pm$0.0)}\\
 & \textbf{$\bm{n}$=1.0} &96\,($\pm$4)&96\,($\pm$4)& 96\,($\pm$4) & \textbf{100\,($\pm$0)} & \textcolor{gray}{100\,($\pm$0)} & 8.0\,($\pm$3.0) & \textbf{7.5\,($\pm$0.5)} & \textcolor{gray}{7.4\,($\pm$0.5)} \\
 & \textbf{$\bm{n}$=2.0} &63\,($\pm$9)& 85\,($\pm$7)&86\,($\pm$7) & \textbf{96\,($\pm$4)} & \textcolor{gray}{98\,($\pm$2)} & 12.2\,($\pm$5.3) & 10.2\,($\pm$1.7) & \textcolor{gray}{9.8\,($\pm$2.0)} \\
 & \textbf{$\bm{n}$=3.0} &31\,($\pm$11)&74\,($\pm$10)& 75\,($\pm$8) & \textbf{86\,($\pm$8)} & \textcolor{gray}{91\,($\pm$7)} & - & 15.6\,($\pm$3.2) & \textcolor{gray}{14.7\,($\pm$2.3)} \\ \midrule
 
\multirow{4}{*}{\textbf{\begin{tabular}[c]{@{}c@{}}Stack in order\\ with noise $\bm{n}$\\ random drop $\bm{p}$=0.1\end{tabular}}} & \textbf{$\bm{n}$=0.0} &71\,($\pm$9)&94\,($\pm$7)& 94\,($\pm$6) & \textbf{98\,($\pm$4)} & \textcolor{gray}{99\,($\pm$1)} & 10.0\,($\pm$3.6) & \textbf{9.4\,($\pm$1.7)} & \textcolor{gray}{9.9\,($\pm$1.9)} \\
 & \textbf{$\bm{n}$=1.0} &71\,($\pm$9)&\textbf{94\,($\pm$7)}& \textbf{94\,($\pm$7)} & \textbf{94\,($\pm$7)}& \textcolor{gray}{97\,($\pm$2)} & 10.7\,($\pm$3.9) & \textbf{10.6\,($\pm$3.2)} & \textcolor{gray}{10.9\,($\pm$3.0)} \\
 & \textbf{$\bm{n}$=2.0} &54\,($\pm$12)&79\,($\pm$9)& 79\,($\pm$8) & \textbf{92\,($\pm$6)} & \textcolor{gray}{95\,($\pm$3)} & - & \textbf{14.5\,($\pm$3.4)} & \textcolor{gray}{15.3\,($\pm$3.5)} \\
 & \textbf{$\bm{n}$=3.0} &21\,($\pm$9)&33\,($\pm$10)& 34\,($\pm$10) & \textbf{55\,($\pm$10)} & \textcolor{gray}{64\,($\pm$8)} & - & - & \textcolor{gray}{-} \\ \bottomrule[1.0pt]
\end{tabular}
% \vspace{-3mm}
\caption{Success rates and task execution time under different degrees of disturbances. We only measure execution time under high success rates. The results show the mean and standard deviation over 4 different seeds, each with 12 episodes.}
\label{arm_table}
\vspace{-2mm}
\end{table*}

\section{Experiments}

In this section, we conduct experiments involving both robotic arm manipulation tasks and humanoid robot tasks, as shown in Figure \ref{exp}. These tasks incorporate various environmental disturbances and imperfect controllers, such as random dropping by the robot end-effector, noise in end-effector placement positions, failure in pick, and unexpected obstacles appearing in the robot's path.

We aim to answer the following questions: (1) Does \textbf{DoReMi} enable immediate detection and recovery from plan-execution misalignment? (2) Does \textbf{DoReMi} lead to higher task success rates and shorter task execution time under environmental disturbances or imperfect controllers?
\begin{figure*}[t]
% \vspace{-1mm}
    \centering
    \includegraphics[width=0.95\linewidth]{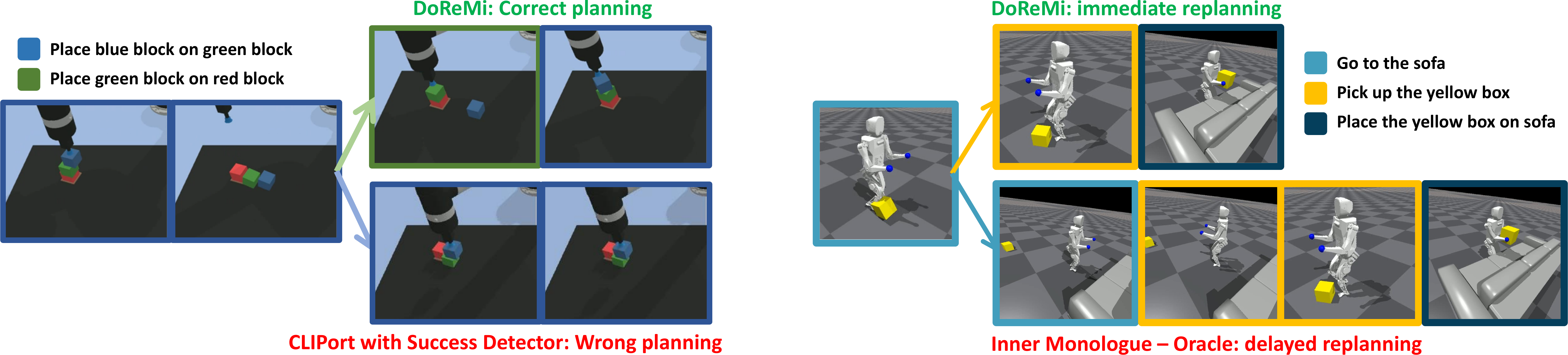}
    % \vspace{-3mm}
    \caption{Comparison examples with baselines. In the left figure, the blocks collapse during manipulation. DoReMi detects this misalignment and replans to pick and place the green block first while the baseline continues to repeat the previous step and results in failure. In the right figure, the box dropped during transportation, DoReMi immediately detects the violation and replan, which is more efficient than baseline.}
    \label{exp1}
    \vspace{-4mm}
    % \vspace{-2mm}
\end{figure*}

\subsection{Robot Arm Manipulation Tasks}

\textbf{Robot and Environment}
This environment is adapted from \textit{Ravens} \cite{zeng2020transporter}, a benchmark for vision-based robotic manipulation focused on pick-and-place tasks. An UR5e robot equipped with a suction gripper operates on a black tabletop, while a third-view camera provides a comprehensive view of the tabletop. The robot possesses a basic skill set including \textit{"pick obj"} and \textit{"place obj on receptacle"}, both of which are pre-trained primitives conditioned on single-step instructions similar to the CLIPort \cite{shridhar2022cliport} and Transporter Nets \cite{zeng2020transporter}. To assess the effectiveness of our algorithm, we introduce additional disturbances into the original environment and the robot controller.

\textbf{Tasks: (1) Pick and Place.}
The agent is required to pick a certain block and place it in a fixture. We assume the block has a probability $p$ to drop every second when sucked by the end-effector, so the agent may need to perform pick and place several times to finish the task.
\textbf{(2) Stack blocks in order.}
The robot is required to stack several blocks in an order given by language instructions. The agent must perform "pick" and "place" skills in a precise sequence to successfully accomplish the task.  We assume the controllers are not perfect by introducing uniform $[0, n]$ cm noise to the place positions. There is also a probability $p$ that a block held by the end-effector might randomly drop every second. The max execution time for all tasks is set to 20 seconds. Any execution that takes time longer than 20 seconds is considered as failure.
% \subsubsection{Experiments Details}

\textbf{Experiment Details} Following the pipeline in Figure \ref{method_new}, we use Vicuna-13B \cite{vicuna2023} as LLM planner and zero-shot transferred BLIP-2 \cite{li2023blip} as VLM constraint detector.
We compare \textbf{DoReMi} with 4 baselines: \textbf{(1) SayCan}: an LLM is utilized to decompose instructions into steps and execute them sequentially. However, this approach assumes the successful execution of each step without considering potential failures. \textbf{(2) CLIPort}: a multi-task CLIPort policy conditioned on the single pick-place step. It utilizes an LLM to decompose instructions into steps and repeat each step until success. The same VLM is leveraged as a success detector to determine whether the current step should be repeated. \textbf{(3) Inner Monologue (IM)}: The same VLM is employed as scene descriptors and success detector to help LLM re-plan upon completion of each step. \textbf{(4) IM-Oracle:} Inner-Monologue with oracle feedback which does not exist in practical real-world settings. Results are shown in Table \ref{arm_table}.

% \subsubsection{Analyses}
\textbf{Result Analyses} \label{analysis1}
In the presence of disturbances, SayCan consistently fails in all tasks due to its lack of success detectors and re-planning mechanisms. In simple pick-place tasks, CLIPort and Inner-monologue with success detector can repeat the step and recover. However, they do not have a mechanism to abort the current execution and only re-plan at the end of each skill, resulting in a longer execution time. In the stack-block task, when encountering situations that require re-planning (e.g., the blocks collapse), CLIPort that only repeats the previous step fails to recover, as shown in Figure \ref{exp1}. 
When provided with imperfect scene descriptors (VLM), Inner Monologue also struggles to recover due to ambiguous open-ended scene descriptions. In contrast, DoReMi leverages LLMs to propose specific constraints for every low-level skill, with the VLM focused on these constraints, leading to highly accurate feedback. Furthermore, our VLM continuously detects constraint violations throughout the execution period, which enables immediate re-planning and recovery. Under these two mechanisms, DoReMi reaches higher success rates and shorter execution times.

\begin{table*}[t]\centering
\small
% \tiny
\begin{tabular}{cl|ccccc|cc}\toprule[1.0pt]
\multicolumn{2}{c|}{\multirow{2}{*}{\textbf{Tasks with disturbance}}} & \multicolumn{5}{c|}{\textbf{Success Rate(\%) \bm{$\uparrow$}}} & \multicolumn{2}{c}{\textbf{Execution Time(s) \bm{$\downarrow$}}} \\ 
\multicolumn{2}{c|}{} & \textbf{\begin{tabular}[c]{@{}c@{}}SayCan\end{tabular}} & \textbf{\begin{tabular}[c]{@{}c@{}}IM\end{tabular}}  &
\textbf{\begin{tabular}[c]{@{}c@{}}DoReMi \\(ours)\end{tabular}} & \textbf{\begin{tabular}[c]{@{}c@{}}DoReMi-FT\\ (ours)\end{tabular}} & \textcolor{gray}{\textbf{\begin{tabular}[c]{@{}c@{}}IM-\\ Oracle \end{tabular}}} & \textbf{\begin{tabular}[c]{@{}c@{}}DoReMi-FT \\(ours)\end{tabular}} & 
\textcolor{gray}{\textbf{\begin{tabular}[c]{@{}c@{}}IM-\\ Oracle\end{tabular}}} \\ \midrule[0.5pt]

\multirow{3}{*}{\textbf{\begin{tabular}[c]{@{}c@{}}Obstacle-avoidance \\ with density $d$ \end{tabular}}} & \textbf{$\bm{d}$=0.0} &\textbf{100\,($\pm$0)}& \textbf{100\,($\pm$0)}&\textbf{100\,($\pm$0)} & \textbf{100\,($\pm$0)} &\textcolor{gray}{100\,($\pm$0)} & \textbf{24.2\,($\pm$0.8)}  &\textcolor{gray}{24.2\,($\pm$0.8)} \\
 & \textbf{$\bm{d}$=0.3} &68\,($\pm$6)&68\,($\pm$6)& \textbf{92\,($\pm$6)} & \textbf{92\,($\pm$6)} & \textcolor{gray}{68\,($\pm$6)} &  \textbf{31.2\,($\pm$2.4)}  & \textcolor{gray}{-} \\
 & \textbf{$\bm{d}$=0.6} & 40\,($\pm$8)&40\,($\pm$8)& \textbf{90\,($\pm$6)} & \textbf{90\,($\pm$6)} & \textcolor{gray}{40\,($\pm$8)} & \textbf{34.3\,($\pm$3.2)}  & \textcolor{gray}{-} \\ \midrule
 
\multirow{3}{*}{\textbf{\begin{tabular}[c]{@{}c@{}}Move-box with\\ random drop $p$\end{tabular}}} & \textbf{$\bm{p}$=0.0} &\textbf{98\,($\pm$2)}&\textbf{98\,($\pm$2)}& \textbf{97\,($\pm$2)} & \textbf{97\,($\pm$2)} & \textcolor{gray}{98\,($\pm$2)}  & \textbf{32.2\,($\pm$2.5)} & \textcolor{gray}{32.1\,($\pm$2.5)}\\
 & \textbf{$\bm{p}$=0.02} &61\,($\pm$7)&63\,($\pm$7)& \textbf{95\,($\pm$4)} & \textbf{96\,($\pm$4)} & \textcolor{gray}{98\,($\pm$2)} & \textbf{35.0\,($\pm$3.0)} & \textcolor{gray}{46.5\,($\pm$4.7)} \\
 & \textbf{$\bm{p}$=0.04} &42\,($\pm$9)& 46\,($\pm$9)&\textbf{94\,($\pm$4)} & \textbf{96\,($\pm$4)} & \textcolor{gray}{96\,($\pm$2)}  & \textbf{37.3\,($\pm$3.1)} & \textcolor{gray}{61.2\,($\pm$7.6)} \\ \midrule
 
\multirow{3}{*}{\textbf{\begin{tabular}[c]{@{}c@{}}Prepare-food with\\ pick failure $p_1$=0.1 \\ random drop $p$ \end{tabular}}} & \textbf{$\bm{p}$=0.0} &78\,($\pm$5) &83\,($\pm$4)& 85\,($\pm$6) & \textbf{96\,($\pm$3)} & \textcolor{gray}{99\,($\pm$1)}  & \textbf{27.6\,($\pm$2.7)} & \textcolor{gray}{27.8\,($\pm$3.0)} \\
 & \textbf{$\bm{p}$=0.02} &49\,($\pm$5)&56\,($\pm$5)& 66\,($\pm$4) & \textbf{93\,($\pm$5)}& \textcolor{gray}{97\,($\pm$2)} &\textbf{31.0\,($\pm$3.8)} & \textcolor{gray}{36.8\,($\pm$5.8)} \\
 & \textbf{$\bm{p}$=0.04} &18\,($\pm$5)&21\,($\pm$7)& 37\,($\pm$8) & \textbf{91\,($\pm$6)} & \textcolor{gray}{96\,($\pm$2)}  & \textbf{35.2\,($\pm$6.5)} & \textcolor{gray}{46.3\,($\pm$7.5)} \\ \midrule
 \textbf{Prepare-food-real} & & -&20&20&\textbf{90}&- &\textbf{195.0}&-  \\ \midrule[0.75pt]
 \textbf{Stack-real} & & -&10&20&\textbf{80}&- &\textbf{240.0}&-  \\

 \bottomrule[1.0pt]
\end{tabular}
% \vspace{-2mm}
\caption{Success rates and task execution time under different degrees of disturbances. We only evaluate execution time under high task success rates.}
\label{hum_table}
\vspace{-3mm}
\end{table*}

% \vspace{-2mm}
% \begin{figure*}[t]
% \begin{minipage}{.5\linewidth}
% \centering
% \includegraphics[width=\linewidth]{figures/pic5_clip_2.pdf}
% \caption{VLM detector fine-tuned on the small dataset can benefit unseen objects, unseen background, and unseen tasks.}\label{ft}
% \end{minipage}
% \hspace{5mm}%
% \begin{minipage}{.48\linewidth}
% \vspace{-1mm}
% \centering
% \includegraphics[trim={0.0cm 8cm 0.0cm 0.0cm},clip,width=1.0\linewidth]{figures/pic6_clip_2.pdf}
% \caption{Box dropped during the execution of skill "Go to the sofa". Inner Monologue only re-plans when the current skill is finished, taking more time to complete the task.}\label{inner_oracle}
% \end{minipage}
% \vspace{-5mm}
% \end{figure*}

\subsection{Humanoid Robot Tasks}
\textbf{Robot Description and Low-level Skill Set}
The humanoid robot utilized in our experiments possesses $6$ degrees of freedom per leg and $4$ degrees of freedom per arm, totaling $20$ degrees of freedom. Controlling complex humanoid robots with a single policy is challenging. Following the framework in \cite{ma2022combining}, we employ reinforcement learning to train the locomotion policy and leverage model-based controllers to acquire the manipulation policy. Specifically, we utilize the Deepmimic algorithm \cite{peng2018deepmimic} to train a locomotion policy conditioned on commanded linear and angular velocity, allowing the robot to execute low-level skills such as "go forward 10 meters," "move forward at speed $v$," "go to \textit{target place}," "turn right/left," and more. As for the manipulation policy, in simulation, we introduce an assistant pick-primitive similar to \cite{li2023behavior}; In the real world, we use dexterous hands with factory-designed pick primitives. These setups allow the robot to execute low-level skills like ``pick up \textit{object}" and ``place \textit{object} on \textit{receptacle}". 

\textbf{Real world robot setup} The real humanoid robot is equipped with the basic low-level controllers listed above. The LLMs and VLMs are running on the cloud and communicate with the robot at 1Hz.
% Detailed architecture and training process can be found in Appendix \ref{Human_task}.

% \subsubsection{Task Categories}

\begin{figure}[t]
\vspace{-1mm}
    % \centering
    \includegraphics[width=\linewidth]{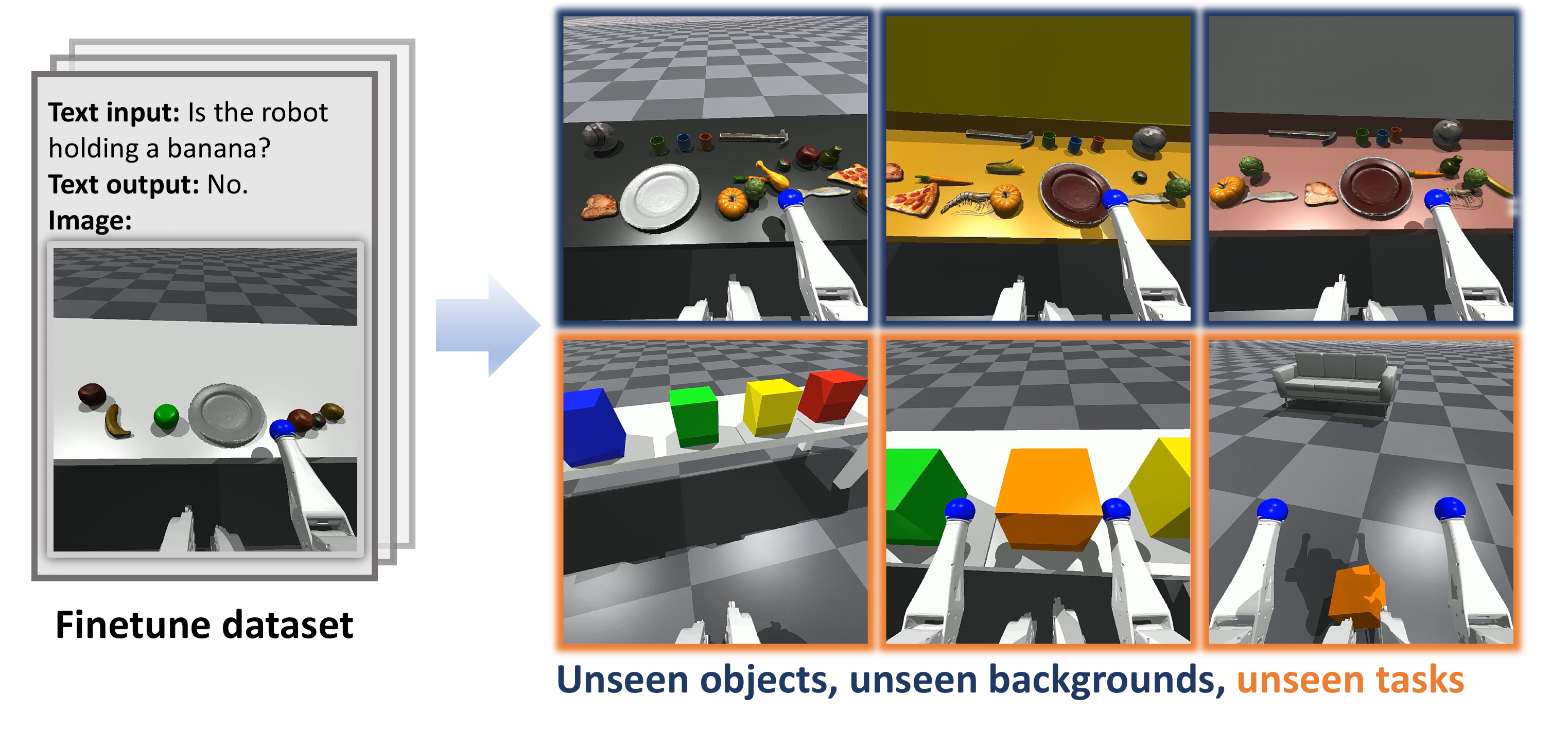}
    % \vspace{-3mm}
\caption{VLM detector fine-tuned on the small dataset can generalize to unseen objects, unseen background, and even unseen tasks.}\label{ft}
    \vspace{-2mm}
\end{figure}

\begin{figure}[t]
\vspace{-1mm}
    % \centering
    \includegraphics[width=\linewidth]{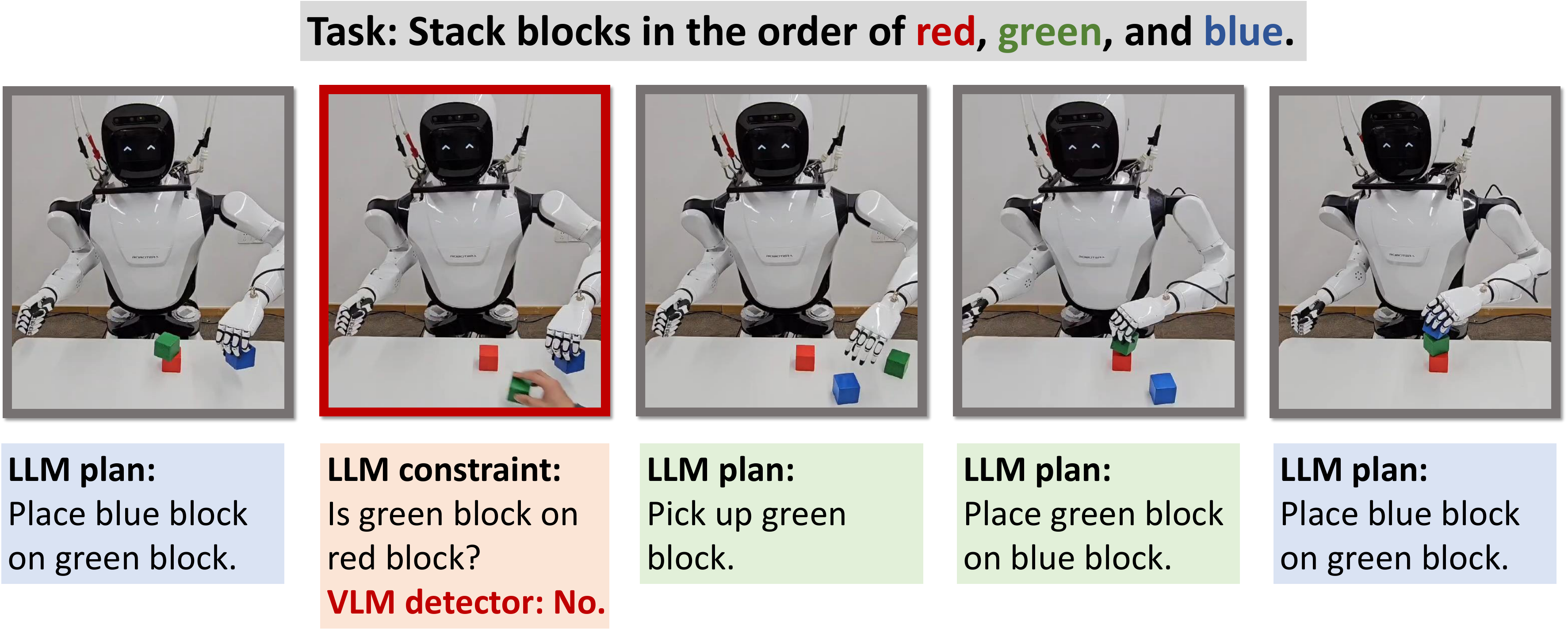}
    \vspace{-3mm}
    \caption{Stacking task on real humanoid robot. During the step ``place blue block on green block", green block was knocked down by external forces. DoReMi-FT successfully recovered from this misalignment }\label{real robot}
    \vspace{-5mm}
\end{figure}

We consider 5 categories of tasks:
\begin{enumerate}
    \item \textbf{Obstacle-avoidance.} The robot is commanded to reach a finish line with unknown obstacles appearing on the way with density $d$. 
    % As we mentioned above, the robot lacks perfect navigation skills and only holds low-level skills such as \textit{"go forward"}, \textit{"turn left/right"}, etc. 
    Therefore, the robot needs to satisfy the constraint \textit{"no obstacle in the front"}. If the constraint is violated, it must perform skill \textit{"turn left/right"} to avoid the collision.
    \item \textbf{ Move-box.} The robot is required to transport a certain box from one location to another. 
A proper solution might involve 1) Go to place A. 2) Pick up box. 3) Go to place B. 4) Put down box. 
We introduced additional perturbations to this task by assuming that the robot has a probability $p$ of dropping the box every second during transport.
\item \textbf{Prepare-food.} The robot is required to collect 2-5 types of foods from random positions according to abstract language instructions with pick failure probability $p_1$ and drop probability $p$ (example in Figure \ref{vlm_compare}b).
\item \textbf{Prepare-food-real} This pick-place experiment is performed on a real humanoid robot. We add external disturbances to knock off the carried object.

\item \textbf{Stack-real} A Real humanoid robot is required to stack the block in a certain order. Blocks may collapse under external forces or place noise. An example is shown in Figure \ref{real robot}.

\end{enumerate}

\textbf{Baselines} Following the pipeline in Figure \ref{method_new}(c), we use Vicuna-13B \cite{vicuna2023} as the LLM planner and BLIP-2 \cite{li2023blip} as the VLM constraint detector. Additionally, we use \textbf{DoReMi-FT} to denote DoReMi with fine-tuned VLM.
We compare our methods with (1) \textbf{SayCan} \cite{ahn2022can}, (2) \textbf{Inner Monologue (IM)} \cite{huang2022inner}, (3) \textbf{IM-Oracle}: Inner monologue with Oracle perfect feedback. Since oracle feedback does not exist in practical real-world settings, we just use this baseline to reflect the upper bound of the planning performance. 

% We compare our methods with (1) \textbf{SayCan} \cite{ahn2022can} which assumes every step is executed successfully, and (2) \textbf{Inner Monologue (IM)} \cite{huang2022inner} which plans at the end of each step and uses the same vision-language model as both success detectors and scene descriptors. \rebuttal{(3) \textbf{Periodic replan} which re-plans at a fixed time interval of 3 seconds and obtains feedback from the same VLM scene descriptors.}
% (4) \textbf{IM-Oracle}. Inner monologue with Oracle feedback which does not exist in practical real-world settings.

\textbf{Result Analyses}
The results are shown in Table \ref{hum_table}. Similar to analysis in section \ref{analysis1}, SayCan failed due to the absence of re-planning mechanisms and Inner-monologue failed because of the ambiguity and the low frequency of the feedback. Furthermore, DoReMi-FT even surpasses IM-oracle in execution time while maintaining similar success rates due to its immediate detection and recovery mechanism, as depicted in the right of Figure \ref{exp1}.
% \rebuttal{Additionally, we find that naively increasing the re-plan frequency (Periodic replan baseline) does not necessarily improve success rates and can even lead to performance degradation. These results can be explained intuitively as follows: without sufficiently precise feedback, the more you re-plan, the more mistakes you may make. Higher frequency is beneficial only with precise enough feedback. These results further highlight the advance of DoReMi which enables more precise feedback, thanks to the seamless cooperation between LLMs and VLMs to propose and detect critical constraints.}

In our experiments, we observed that the performance of zero-shot transferred VLM diminishes as the scene complexity increases, such as in the prepare-food(-real) task involving multiple objects. In order to enhance the performance in complex scenarios, we collected a small dataset to fine-tune the pre-trained BLIP-2 VLM \cite{li2023blip}. The dataset only consisted of 128 image-text pairs with 5 demonstrations on fruit objects. In both simulated and real-world experiments, we are delighted to find that DoReMi-FT with the fine-tuned BLIP-2 model can generalize to complicated scenes with unseen objects, unseen backgrounds, and even unseen tasks. As shown in Figure \ref{ft}, test tasks include new categories of objects like junk food, vegetables, and seafood with random positions, as well as unseen backgrounds. Detailed information on the finetune process can be found in the Appendix.

\section{Conclusion}
% \vspace{-2mm}
% \textbf{Limitation}
% Our experiments indicate that the zero-shot transferred VLM is not a perfect constraint detector. We need to fine-tune the VLM in complicated tasks to improve detection accuracy and our framework can benefit from more advanced VLMs in the future. Furthermore, a detector fully based on vision may be limited by mis-detection, occlusion, and perspective. We may explore detectors in other modalities under our framework in the future.

% \textbf{Conclusion} 
When employing language models for embodied tasks in a hierarchical approach, the low-level execution might deviate from the high-level plan. We emphasized the importance of continuously aligning the plan with execution and leveraged LLM to generate both plan and constraints, which enables grounding language through immediate detection and recovery under practical VLM constraint detectors. Variety of challenging tasks in disturbed environments demonstrated the effectiveness of DoReMi.